\documentclass[a4paper,fleqn]{cas-sc}

\usepackage[authoryear,longnamesfirst]{natbib}
\usepackage{array}
\usepackage{multirow}
\usepackage{xcolor}
\usepackage{soul}
\usepackage{comment}
\usepackage{natbib}
\usepackage{totcount}

\newcommand\MyBox[2]{
  \fbox{\lower0.75cm
    \vbox to 1.7cm{\vfil
      \hbox to 1.7cm{\hfil\parbox{1.4cm}{#1\\#2}\hfil}
      \vfil}%
  }%
 }

\def\tsc#1{\csdef{#1}{\textsc{\lowercase{#1}}\xspace}}
\tsc{WGM}
\tsc{QE}
\tsc{EP}
\tsc{PMS}
\tsc{BEC}
\tsc{DE}

\newif\ifshowchanges
\showchangesfalse

\ifshowchanges
    \usepackage{changes} 
\else 
    \newcommand{\added}[1]{#1}
    \newcommand{\deleted}[1]{}
    \newcommand{\replaced}[2]{#1}
\fi

\begin{document}
\let\WriteBookmarks\relax
\def\floatpagepagefraction{1}
\def\textpagefraction{.001}


\shortauthors{Anubrata Das et~al.}

\title [mode = title] {The State of Human-centered NLP Technology for Fact-checking}

\shorttitle{The State of Human-centered NLP Technology for Fact-checking}




%

\author{Anubrata Das}[orcid=0000-0002-5412-6149]

\cormark[1]





\affiliation[1]{
    organization={School of Information, The University of Texas at Austin}, 
    city={Austin},
    state={TX},
    country={USA}}

\author{Houjiang Liu}[orcid=0000-0003-0983-6202]

\author{Venelin Kovatchev}[orcid=0000-0003-1259-1541]

\author{Matthew Lease}[orcid=0000-0002-0056-2834]



\cortext[cor1]{Corresponding author}



\begin{abstract}
Misinformation threatens modern society by promoting distrust in science, changing narratives in public health, heightening social polarization, and disrupting democratic elections and financial markets, among a myriad of other societal harms. To address this, a growing cadre of professional fact-checkers and journalists provide high-quality investigations into purported facts. However, these largely manual efforts have struggled to match the enormous scale of the problem. In response, a growing body of Natural Language Processing (NLP) technologies have been proposed for more scalable fact-checking. Despite tremendous growth in such research, however, practical adoption of NLP technologies for fact-checking still remains in its infancy today.

In this work, we review the capabilities and limitations of the current NLP technologies for fact-checking. Our particular focus is to further chart the design space for how these technologies can be harnessed and refined in order to better meet the needs of human fact-checkers. To do so, we review key aspects of NLP-based fact-checking: task formulation, dataset construction, modeling, and human-centered strategies, such as explainable models and human-in-the-loop approaches. Next, we review the efficacy of applying NLP-based fact-checking tools to assist human fact-checkers. We recommend that future research include collaboration with fact-checker stakeholders early on in NLP research, as well as incorporation of human-centered design practices in model development, in order to further guide technology development for human use and practical adoption. Finally, we advocate for more research on benchmark development supporting extrinsic evaluation of human-centered fact-checking technologies.

\end{abstract}




\maketitle

\section{Introduction}





Misinformation and related issues (disinformation, deceptive news, clickbait, rumours, and information credibility) increasingly threaten society.
While concerns of misinformation existed since the early days of written text \citep{marcus1992mesoamerican}, with recent development of social media, \added{the }entry barrier for creating and spreading content has never been lower.  \added{Moreover, polarization online drives the spread of misinformation that in turn increases polarization \citep{cinelli2021dynamics,cinelli2021online,vicario2019polarization}. Braking such a vicious cycle would require addressing the problem of misinformation at its root.}

Fields
such as journalism \citep{graves2018boundaries, graves2019fact, neely2018focus} and archival studies \citep{lebeau2017entitled} have extensively studied misinformation, and recent years have seen a significant growth in fact-checking initiatives \added{to address this problem}. Various organizations now focus on fact-checks (e.g., PolitiFact, 
Snopes, 
FactCheck, 
First Draft, 
and Full Fact), and organizations such as the International Fact-Checking Network (IFCN)\footnote{\url{https://www.politifact.com/},\, \url{https://www.snopes.com/},\,  \url{https://www.factcheck.org/},\, \url{https://firstdraftnews.org/},\, \url{https://fullfact.org/},\, and \url{https://www.poynter.org/ifcn/}, respectively.} train and provide resources for independent fact-checkers and journalists to further scale expert fact-checking.

While professional fact-checkers and journalists provide high-quality investigations of 
purported facts to inform the public, human effort struggles to match the global Internet scale of the problem. 
To address this, a growing body of research has investigated Natural Language Processing (NLP) to \added{fully or partially} automate fact-checking \citep{Guo2021ASO, Nakov2021AutomatedFF, zhou2020survey, zeng2021automated, graves2018understanding}. 
However, even state-of-the-art NLP technologies still cannot match human capabilities in many areas and 
remain insufficient to automate fact-checking in practice. Experts argue \citep{ArnoldPh:online,Nakov2021AutomatedFF} that fact-checking is a complex process and requires subjective judgement and expertise. While current NLP systems are increasingly better at addressing simple fact-checking tasks, identifying false claims that are contextual and beyond simple declarative statements \replaced{remains}{are yet} beyond the reach for fully automated systems \citep{chen2022generating,fan2020generating}.
For example, claims buried in conversational systems, comment threads in social media community, and claims in multimedia contents are particularly challenging for automated systems. Additionally, \added{most fact-checking practitioners} desire NLP tools that are integrated into the existing fact-checking workflow and \deleted{assist in} \replaced{reduce}{reducing} latency \deleted{are most desired by fact-checking practitioners }\citep{Nakov2021AutomatedFF, graves2018boundaries, alam2020fighting}.

In this literature review, \added{we provide the reader with a comprehensive and holistic overview of the} \deleted{we first examine the} current state-of-the-art challenges and opportunities to more effectively leverage NLP technology in fact-checking. 
\added{Our objectives in this work are twofold. First, we cover all aspects of the NLP pipeline for fact checking: }
\deleted{We investigate} task formulation, dataset construction, and modeling approaches. 
\added{Second, we emphasize the human-centered approaches that seek to augment and accelerate human fact-checking, rather than supplant it.
In contrast,}
\deleted{While} prior literature reviews \citep{zeng2021automated,Guo2021ASO, Oshikawa2020ASO} \added{either} provide an overview of the existing approaches or capture the details of \added{only a }specific part of the fact-checking pipeline \citep{Kotonya2020ExplainableAF, Hardalov2021ASO, Hanselowski2018ARA, demartini2020human}. 
\deleted{, our review emphasizes human-centered approaches that seek to augment and accelerate human fact-checking,} 
\deleted{rather than supplant it.}

Furthermore, we argue that it is important to extend the review of NLP technologies for fact-checking from modeling development to the area of Human-Computer Interaction (HCI) because technology design should reflect user needs so that its development can be better integrated in the real-world use context \citep{graves2018understanding, lease2020designing, kovatchev2020your, Nakov2021AutomatedFF, Micallef2022, juneja2022human}. \added{Specifically, we point the reader towards Section \ref{sec:future} where we propose concrete directions for future work.}

Current challenges are largely due to the relatively early stage of development of the automated fact-checking technology.
Specifically, current studies tend to adopt an intrinsic evaluation of components of the fact-checking pipeline rather than an end-to-end extrinsic evaluation of the entire fact-checking task. Moreover, component-wise accuracies may remain below the threshold required for practical adoption. Furthermore, while the research community’s focus on prediction accuracy has yielded laudable improvements, human factors (e.g., usability, intelligibility, trust) have garnered far less attention or progress yet are crucial for practical adoption.

Such limitations have implications for future research. First, practical use of NLP technologies for fact-checking is likely to come from hybrid, human-in-the-loop approaches rather than full automation. Second, as the technology matures, end-to-end evaluation becomes increasingly important to ensure practical solutions are being developed to solve the real-world use-case. To this end, new benchmarks that facilitate the extrinsic evaluation of automated fact-checking applications in practical settings may help drive progress on solutions that can be adopted for use in the wild.
Finally,  
to craft effective human-in-the-loop systems, more cross-cutting NLP and HCI integration could strengthen design of 
fact-checking tools, so that they are accurate, scalable, and usable in practice. Toward this end, it may be fruitful to collaborate more with stakeholders early on in NLP research and incorporate human-centered design practices in developing models.

\added{We have written this article with different audiences in mind. For researchers and fact-checkers who are new to automated fact-checking, this article provides a comprehensive overview of the problem. We discuss the challenges, the state-of-the-art capabilities, and the opportunities in the field, and we emphasize how machine learning and natural language processing can be used to combat disinformation. We recommend researchers new to this topic read the article in its entirety, following the logical structure of sections. Other readers who have more experience in the field may already be familiar with some of the concepts that we discuss. For them, \textit{this paper offers a novel human-centered perspective of automated fact checking} and a discussion on how that perspective can affect system design, implementation, and evaluation. To facilitate the use of the paper by more experienced readers, we provide a quick overview of the content covered in each section.}




\begin{itemize}
    \item \added{Section \ref{sec:pipeline} introduces the automated fact-checking pipeline. We provide an overview of the process for human fact-checkers and for automated solutions.}
    \item \added{Section \ref{sec:problem-formulation} discusses the \textit{task formulation}: the goals and formal definitions of different sub-tasks in fact checking.} 
    \item \added{Section \ref{sec:datasets} describes the process of dataset construction, presents the most popular corpora for automated fact checking, and outlines some limitations of the data.}
    \item \added{Section \ref{sec:approaches} reviews approaches for automating fact checking. We discuss general NLP capabilities (Section \ref{subsec:nlp}), explainable approaches (Section \ref{subsec:explainable}), and human-in-the-loop (Section \ref{subsec:hitl}) approaches for fact-checking.}
    \item \added{Section \ref{sec:tools} surveys existing tools that apply NLP for fact-checking in a practical, real-world context. We argue that the human-centered perspective is necessary for the practical adoption of automated solutions.}
    \item \added{Section \ref{sec:future} provides future research directions in the context of human-centered fact checking. We discuss the work division between human and AI for mixed-initiative fact-checking in Section \ref{subsec:articulation}. In Section \ref{subsec:benchmark} we propose a novel concept for measuring trust and a novel human-centered evaluation of NLP to assist fact-checkers.}
    \item \added{We conclude our literature review with Section \ref{sec:con}}.
\end{itemize}

\deleted{
The rest of this article is organized as follows. 
Section \ref{sec:pipeline} introduces the automated fact-checking pipeline. Sections \ref{sec:problem-formulation} and \ref{sec:datasets} respectively discuss task formulation and dataset construction for automated fact-checking. Section \ref{sec:approaches} discusses general NLP capabilities (Section \ref{subsec:nlp}), explainable approaches (Section \ref{subsec:explainable}), and human-in-the-loop (Section \ref{subsec:hitl}) approaches for fact-checking. Section \ref{sec:tools} surveys existing tools that apply NLP for fact-checking in a practical, real-world context. Section \ref{sec:future} proposes future research directions, including articulating work division between human and AI for mixed-initiative fact-checking (Section \ref{subsec:articulation}) and conducting human-centered evaluation of NLP to assist fact-checkers (Section  \ref{subsec:benchmark}). We conclude the review in Section \ref{sec:con}.
}

\begin{figure}
    \centering
    \includegraphics[scale=0.5]{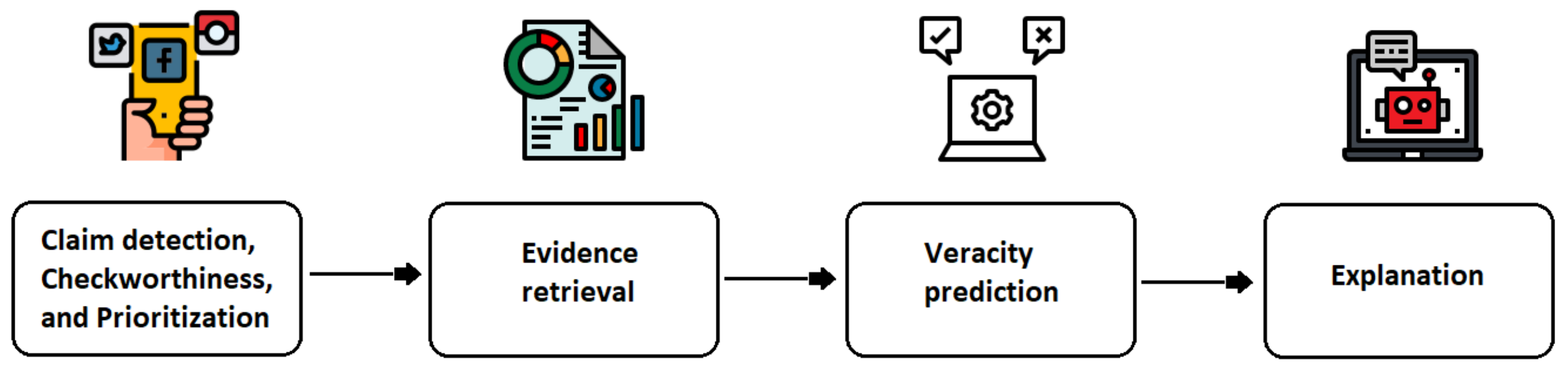}
    \caption{Fact-checking pipeline}
    \label{fig:guo-fact-checking}
\end{figure}

\section{Fact-Checking Pipeline} \label{sec:pipeline}

The core idea behind automated fact-checking is enabling AI to reason over available information to determine the truthfulness of a claim.
For successful automation, it is essential first to understand the complex process of journalistic fact-checking that involves human expertise along with skilled effort towards gathering evidence and
synthesizing the evidence. Additional complexity comes from the need to process heterogeneous sources (e.g., information across various digital and non-digital sources). Data is also spread across different modalities such as images, videos, tables, graphs, among others. Moreover, there is a lack of tools that support effective and efficient fact-checking workflows \citep{graves2018understanding,Nakov2021AutomatedFF,ArnoldPh:online}. 

\citet{graves2017anatomy} breaks down the practical fact-checking mechanism for human fact-checkers into multiple steps such as 
a)~identifying the claims to check, b)~tracing false claims, c)~consulting experts, and d)~sharing the resulting fact-check. A growing body of AI literature --- specifically in NLP --- focuses on automating the fact-checking process. 
We synthesize several related surveys \citep{Guo2021ASO,Nakov2021AutomatedFF,graves2018understanding,zeng2021automated, Micallef2022} and distinguish four typical stages 
that constitute the automated fact-checking technology pipeline (illustrated in Figure \ref{fig:guo-fact-checking}). 
Note that the pipeline we describe below closely follows the structure of \citet{Guo2021ASO}, though the broader literature is also incorporated within these four stages: 
\begin{itemize}
    \item {\bf Claim Detection, Checkworthiness, and Prioritization:} Claim detection involves monitoring news and/or online information for potentially false content to fact-check. One must identify claims that are potentially falsifiable (e.g., purported facts rather subjective opinions) \citep{Guo2021ASO,zeng2021automated}. Moreover, because it is impractical to fact-check everything online given limited fact-checking resources (human or automated), fact checkers must prioritize what to fact-check \citep{ArnoldPh:online}. NLP researchers have sought to inform such prioritization by automatically predicting the "checkworthiness" of claims \citep{Nakov2021AutomatedFF}. \added{Additionally, to avoid repeated work, fact-checkers may consult existing fact-checking databases before judging the veracity of a new claim ({\em claim matching \citep{zeng2021automated}}). We see claim matching as a part of prioritizing claims, as fact-checkers would prioritize against checking such claims.} 
    \item {\bf Evidence Retrieval:} Once it is clear which claims to fact-check, the next step is to gather relevant, trustworthy evidence for assessing the claim \citep{Guo2021ASO,zeng2021automated}. 
    \item {\bf Veracity Prediction:} Given the evidence, it is  necessary to assess it to determine the veracity of the claim \citep{Guo2021ASO,zeng2021automated}. 
    \item {\bf Explanation:} Finally, for human use, one must explain the fact-checking outcome via human-understandable justification for the model's determination \citep{graves2018understanding,Kotonya2020ExplainableAF,Guo2021ASO}. 
\end{itemize}

In the subsequent sections, we discuss each of the tasks above in the context of existing NLP research in automated fact-checking. Some other steps (for example, detecting propaganda in text, click-bait detection) are also pertinent to fact-checking but do not directly fit into the stages described above. They are briefly discussed in Section \ref{subsec:task-other}. 

\section{Task Formulation for Automated Fact-Checking} \label{sec:problem-formulation}

\replaced{Modern Natural Language Processing is largely-data driven.}{Modern machine learning, including NLP, is largely data-driven.}
In this article, we distinguish task formulation (conceptual) vs. dataset construction (implementation activity, given the task definition). That said, the availability of a suitable dataset or the feasibility of constructing a new dataset can also bear on how tasks are formulated. 


\subsection{Claim Detection, Checkworthiness, and Prioritization} \label{subsec:task-claim}
Fact-checkers and news organizations monitor information sources such as social media (Facebook, Twitter, Reddit, etc.), political campaigns and speeches, and public addresses from government officials on critical issues \citep{ArnoldPh:online, Nakov2021AutomatedFF}. Additional sources include tip-lines on end-to-end encrypted platforms (such as WhatsApp, Telegram, and Signal) \citep{Kazemi2021TiplinesTC}. 
The volume of information on various platforms makes it challenging to efficiently monitor all sources for misinformation. 
\citet{zeng2021automated} define the {\em claim detection} step as identifying, filtering, and prioritizing claims. 

To identify claims, social media streams are often monitored for rumors \citep{Guo2021ASO}. A rumour can be defined as a claim that is unverified and being circulated online \citep{zubiaga2018detection}. Rumours are characterized by the subjectivity of the language and the reach of the content to the users \citep{qazvinian2011rumor}. Additionally, metadata related to virality, such as the number of shares (or retweets and re-posts), likes, or comments are also considered when identifying whether a post is a rumour \citep{zhang2021mining,gorrell2019semeval}. However, detecting rumours alone is not sufficient to decide whether a claim needs to be fact-checked.

For each text of interest, the key questions fact-checking systems need to address include:
\begin{enumerate}
    \item ~Is there a claim to check?
    \item ~Does the claim contain verifiable information?
    \item ~Is the claim checkworthy?
    \item ~Has a trusted source already fact-checked the claim?
\end{enumerate}

Regarding the first criterion --- is there a claim --- one might ask whether the claim contains a purported fact or an opinion \citep{hassan2017toward}. For example, a statement such as \textit{``reggae is the most soulful genre of music''} represents personal preference that is not checkable. In contrast, \textit{``<NAME> won a gold medal in the Olympics''} is checkable by matching <NAME> to the list of all gold medal winners. 

Whether the claim contains verifiable information is more challenging. For example, if a claim can only be verified by private knowledge or personal experience that is not broadly accessible, then it cannot be checked \citep{konstantinovskiy2021toward}. For example, if someone claims to have eaten a certain food yesterday, it is probably impossible to verify beyond their personal testimony. 

As this example suggests, the question of whether the claim contains verifiable information depends in large part on what evidence is available for verification. This, in turn, may not be clear until after evidence retrieval is performed. \replaced{In practice fact-checkers may perform some preliminary research, but mostly try to gauge checkworthiness only based on the claim itself.}{In practice, fact-checkers inspect a claim only based on the text to assess how likely it is to be verifiable.}


In addition, this consideration is only one of many that factors into deciding whether to check a claim.  Even a claim that may appear to be unverifiable may still be of such great public interest that it is worth conducting the fact-check. Moreover, even if the fact-check is conducted and ultimately indeterminate (i.e., evidence does not exist either to verify or refute the claim), simply showing that a claim's veracity cannot be determined may still be a valuable outcome. 


A claim is deemed {\em checkworthy} if a claim is of significant public interest or has the potential to cause harm \citep{nakov2021clef,hassan2015detecting}. For example, a claim related to the effect of a vaccine on the COVID-19 infection rate is more relevant to the public interest and hence more checkworthy than a claim about some philosopher’s favorite food. 

Claims, like memes, often appear several times and/or across multiple platforms (in the same form or with slight modification) \citep{leskovec2009meme,Nakov2021AutomatedFF,ArnoldPh:online}.
Fact-checking organizations maintain a growing database claims which have already been fact-checked. Thus, detected claims are compared against databases of already fact-checked claims by trusted organizations \citep{Shaar2020ThatIA,shaar2021assisting}. Comparing new claims against such databases helps to avoid duplicating work on previously fact-checked claims. This step is also known as {\em claim matching} \citep{zeng2021automated}.

Reports from practitioners argue that if a claim is not checked within the first few hours, a late fact-check does not have much impact on changing the ongoing misinformation narrative \citep{Nakov2021AutomatedFF,ArnoldPh:online}. Moreover, limited resources for fact-checking make it crucial for organizations to prioritize the claims to be checked \citep{borel2016chicago}. 
Claims can be prioritized based on their 
checkworthiness \citep{nakov2022proposal,nakov2021clef}. \citet{nakov2022proposal} note that checkworthiness is determined based on factors such as 
\begin{enumerate} \label{list:checkworthiness}
    \item How urgently a claim needs to be checked?
    \item How much harm can a claim cause \citep{alam2020fighting,Alam2021FightingTC,Shaar2021OverviewOT}?
    \item Would the claim require attention from policy makers for addressing the underlying issue?
\end{enumerate}
Note that estimating harms is quite challenging, especially without first having 
a thorough understanding and measures of harm caused by misinformation\cite{Neumann-2022}. 

The spread of a claim on social media provides another potential signal for identifying public interest \citep{ArnoldPh:online}. 
In the spirit of doing ``the greatest good for the greatest number'', viral claims might be prioritized highly because any false information in them has the potential to negatively impact a large number of people. On the other hand, since fairness considerations motivate equal protections for all people, we cannot serve only the majority at the expense of minority groups \citep{ekstrand2022fairness,Neumann-2022}. Moreover, such minority groups may be more vulnerable, motivating greater protections, and may be disproportionately impacted by mis/disinformation \citep{Guo2021ASO}. See  
Section \ref{subsec:gap-task} for additional discussion. 



\subsection{Evidence Retrieval } \label{subsec:task-evidence}
\added{Some sub-tasks in automated fact-checking can be performed without the presence of explicit evidence. For example, the linguistic properties of the text can be used to determine whether it is machine-generated \citep{Wang2017LiarLP,rashkin2017truth}. However, assessing assessing claim veracity without evidence is clearly more challenging \citep{schuster2020limitations}.}
\deleted{While some works in automated fact-checking have sought to assess claim veracity using only linguistic features indicative of unreliable content \citep{Wang2017LiarLP,rashkin2017truth}, assessing claim veracity without evidence is clearly more challenging \citep{schuster2020limitations}. Additionally, with the recent advances in natural language generation \citep{zellers2020defending}} 
\deleted{it has become more difficult to distinguish deceptive content based on linguistic features alone. }

Provenance of a claim can also signal information quality; known unreliable source or distribution channels are often repeat offenders in spreading false information\footnote{\url{https://disinformationindex.org/}}. Such analysis of provenance can be further complicated when content is systematically propagated by multiple sources (twitter misinformation bots) \citep{jones2019gulf}.

It is typically assumed that fact-checking requires gathering of reliable and trustworthy evidence that provides information to reason about the claim \citep{graves2018understanding,li2016survey}. In some cases, multiple aspects of a claim needs to be checked.
A fact-checker would then decompose such a claim into distinct questions and gather relevant evidence for the question \citep{borel2016chicago,chen2022generating}. From an information retrieval (IR) perspective, we can conceptualize each of those questions as an ``information need'' for which the fact-checker must formulate one or more queries to a search engine \citep{bendersky2012effective} in order to retrieve necessary evidence.

Evidence can be found across many modalities, including text, tables, knowledge graphs, images, and videos. Various metadata can also provide evidence and are sometimes required to assess the claim. Examples include context needed to disambiguate claim terms, or  background of the individual or organization from whom the claim originated. 

Retrieving relevant evidence also depends on the following questions \citep{singh2021case}:
\begin{enumerate}
    \item Is there sufficient evidence available related to a claim?
    \item Is it accessible or available in the public domain? 
    \item Is it in a format that can be read and processed? 
\end{enumerate}
As noted earlier in Section \ref{subsec:task-claim}, the preceding claim detection task involves assessing whether a claim contains verifiable information; this depends in part on what evidence exists to be retrieved, which is not actually known until evidence retrieval is performed. Having now reached this evidence retrieval step, we indeed discover whether sufficient evidence exists to support or refute the claim.


Additionally, evidence should be trustworthy, reputable \citep{Nguyen18-aaai,lease2018fact}, and unbiased \citep{chen2019seeing}. 

Once evidence is retrieved, 
{\em stance detection} assesses the degree to which the evidence supports or refutes the claim \citep{nguyen2018believe,Ferreira2016EmergentAN,popat2018declare}. Stance detection is typically formulated as a classification task (or ordinal regression) over each piece of retrieved evidence. Note that some works formulate stance detection as an independent task \citep{Hanselowski2018ARA,Hardalov2021ASO}. 

\subsection{Veracity Prediction} \label{subsec:task-veracity}

Given a claim and gathered evidence, {\em veracity prediction} involves reasoning over the collected evidence and the claim. 
Veracity prediction can be formulated as a binary classification task (i.e., true vs.\ false) \citep{popat2018declare,nakashole2014language,potthast2018stylometric}, or as a fine-grained, multi-class task following the journalistic fact-checking practices \citep{Augenstein2019MultiFCAR,Shu2020FakeNewsNetAD,Wang2017LiarLP}. In some cases, there may not be enough information available to determine the veracity of a claim \citep{thorne2018fever}. 

Note that fact-checking is potentially a recursive process because retrieved evidence may itself need to be fact-checked before it can be trusted and acted upon \citep{graves2018understanding}. This is also consistent with broader educational practices in information literacy\footnote{\url{https://en.wikipedia.org/wiki/Information_literacy}} in which readers are similarly encouraged to evaluate the quality of information they consume. Such assessment of information reliability can naturally integrate with the 
veracity prediction task in factoring in the reliability of the evidence along with its stance \citep{Nguyen18-aaai, Guo2021ASO}. 

\subsection{Explaining Veracity Prediction} \label{subsec:task-explanation}
While a social media platform might use automated veracity predictions in deciding whether to automatically block or demote content, the use of fact-checking technology often involves a human-in-the-loop, whether it is a platform moderator, a journalist, or an end-user. When we consider such human-centered use of fact-checking technologies, providing an automated veracity prediction without justifying the answer can cause a system to be ignored or distrusted, or even reinforce mistaken human beliefs in false claims (the ``backfire effect'' \citep{lewandowsky2012misinformation}). 
Explanations and justifications are especially important given the noticeable drop in performance of state-of-the-art NLP systems when facing adversarial examples \citep{kovatchev2022longhorns}.
Consequently, automated fact-checking systems intended for human-consumption should seek to explain their veracity predictions in a similar manner to that of existing journalistic fact-checking practices \citep{uscinski2015epistemology}. A brief point to make is that much of the explanation research has focused on explanations for researchers and engineers engaged in system development (types of explanations, methods of generating them, and evaluation regimens). In contrast,we emphasize here explanations for \textit{system users}.

Various types of explanations can be provided, such as through 
\begin{enumerate}
    \item ~evidence attribution
    \item ~explaining the decision-making process for a fact-check
    \item ~summarizing the evidence
    \item ~case-based explanations
\end{enumerate}
{\em Evidence attribution} is the process of identifying evidence or a specific aspect of the evidence (such as paragraphs, sentences, or even tokens of interest) \citep{thorne2018fever,popat2018declare,shu2019defend,lu2020gcan}. Furthermore, the relative importance of the evidence can also justify the fact-checking outcome \citep{nguyen2018believe}. Alternatively, a set of rules or interactions to break down parts of the decision-making process can also serve as an explanation \citep{gad2019exfakt,Nguyen18-aaai}. Such formulation focuses more on how the evidence is processed to arrive at a decision. Explaining the veracity can also be formulated as a summarization problem over the gathered evidence to explain a fact-check \citep{Atanasova2020GeneratingFC,Kotonya2020ExPubhealth}. Finally, case-based explanations can provide the user with similar, human-labeled instances \citep{das2022prototex}.
%




\subsection{\replaced{Related}{Other} Tasks} \label{subsec:task-other}
\added{In addition to tasks that are considered central to the automated fact-checking pipeline, some additional tasks bear mentioning as related and complementary to the fact-checking enterprise. Examples of such tasks} 
\deleted{Additional tasks discussed in the literature are related to the fact-checking pipeline but do not directly fit into the stages described above. Such tasks}
include propaganda detection \citep{Martino2020ASO}, clickbait detection \citep{potthast2016clickbait}, and argument mining \citep{lawrence2020argument}. Furthermore, some tasks can be formulated independent of the fact-checking pipeline and utilized later to improve individual \added{fact-checking} sub-tasks. For example, predicting the virality
of social media content \citep{jain2021reconstructing} can help improve claim detection and claim checkworthiness. Similarly, network analysis on fake news propagation \citep{shao2016hoaxy} can help in analyzing provenance.

With an eye toward building more human-centered AI approaches, \replaced{there are also some}{a few additional} tasks \replaced{that could be applied to}{ that can also} help automate parts of the fact-checking process. \replaced{For example, claim detection might be improved via an}{ A} URL recommendation engine for \replaced{content that}{ fact-checkers on what} might need fact-checking \deleted{on social media sites may improve claim detection} \citep{vo2018rise}. Additionally, fact-checkers could benefit from a predicted score for claim difficulty \citep{singh2021case}. In terms of evidence retrieval and veracity prediction, one might generate fact-checking briefs to aid inexpert fact-checkers \citep{fan2020generating}. Instead of summarizing the evidence in general  (Section \ref{subsec:task-explanation}), one might instead summarize with the specific goal of decision support \citep{hsu2021decision}. 

\subsection{Key Challenges} \label{subsec:gap-task}

Most work in automated fact-checking has been done on veracity prediction, and to a lesser extent, on explanation generation. Recently, we have seen more attention directed towards claim detection and checkworthiness. In contrast, work on evidence retrieval remains less developed.

\paragraph{Claim Detection} 
\citet{Guo2021ASO} points out several sources of biases in the claim check-worthiness task. Claims could be of variable interest to different social groups. Additionally, claims that might cause more harm to marginalized groups compared to the general population may not get enough attention. Ideally, models identifying check-worthiness need to overcome any possible disparate impact. 

Similar concerns appear in the report by Full Fact \citep{ArnoldPh:online}. One of the criteria for selecting a claim for fact-checking across several organizations is “Could the claim threaten democratic processes or minority groups?” However, such criterion may be at odds with
the concerns of virality. Fact-checking organizations often monitor virality metrics to decide which claims to fact-check \citep{ArnoldPh:online,Nakov2021AutomatedFF}. Nevertheless, if a false claim is targeted towards an ethnic minority, such claims may not cross the virality thresholds. 

Prioritizing which claims to fact-check requires attention to various demographic traits: content creators, readers, and subject matter. 
Claim check-worthiness dataset design can thus benefit from 
consideration of demographics.



\paragraph{Evidence Retrieval} Evidence retrieval has been largely neglected in the automated fact-checking NLP literature. It is often assumed that evidence is already available, or, coarse-grained evidence is gathered from putting the claims into a search engine \citep{popat2018declare,nguyen2018believe}. \added{However, \cite{hasanain2022studying} show in their study that search engines optimized for relevance seldom retrieve evidence most useful for veracity prediction.} Although retrieving credible information has been studied thoroughly in IR \citep{clarke2020overview},  more work is needed that is focused on retrieving evidence for veracity assessment \citep{lease2018fact,TRECHeal37}.

\paragraph{Veracity Prediction and Explanation} A critical challenge for automated systems is to reason over multiple sources of evidence while also taking source reputation into account. Additionally, explaining a complex reasoning process is a non-trivial task. The notion of model explanations itself is polysemous and evolving in general, not to mention in the context of fact checking. As explainable NLP develops, automated fact-checking tasks also need to evolve and provide explanations that are accessible to human stakeholders yet faithful to the underlying model. For example, case-based explanations are mostly unexplored in automated fact-checking, although working systems have been proposed for propaganda detection \citep{das2022prototex}.


\noindent \\ \replaced{In many NLP tasks, such as machine translation or natural language inference, the goal is to build fully-automated, end-to-end solutions. However, in the context of fact-checking, }{While a laudable goal of automated fact-checking is to build fully-automated, end-to-end solutions,} state-of-the-art limitations suggest the need for humans-in-the-loop for the forseable future. Given this, automated tooling to support human fact-checkers is crucial. However, understanding the fact-checker needs and 
incorporating those needs in the task formulation has been largely absent from the automated fact-checking literature, with a few notable exceptions \citep{Nakov2021AutomatedFF,demartini2020human}. Future research could benefit from greater involvement of fact-checkers in the NLP research process and shifting goals from complete automation toward human support. 

\section{Dataset Construction} \label{sec:datasets}

Corresponding to task formulation (Section \ref{sec:problem-formulation}), our presentation of fact-checking datasets is also organized around claims, evidence, veracity prediction, and explanation. Note that not all datasets have all of these components. 


\subsection{Claim detection and claim check-worthiness}


Claim detection datasets typically contain claims and their sources (documents, social media streams, transcripts from political speeches) \citep{Guo2021ASO}. One form of claim detection is identifying rumours on social media, where datasets are primarily constructed with text from Twitter \citep{zubiaga2018detection,qazvinian2011rumor} and Reddit \citep{gorrell2019semeval,lillie2019joint}. Some works provide the claims in the context they appeared on social media \citep{zhang2021mining,Ma2016DetectingRF}. However, several studies note that most claim detection datasets do not contain enough context. As the discussion of metadata in Section \ref{sec:problem-formulation} suggests, broader context might include: social media reach, virality metrics, the origin of a claim, and relevant user data (i.e., who posted a claim, how influential they are online, etc.) 
\citep{ArnoldPh:online,Nakov2021AutomatedFF}. 

Claim check-worthiness datasets \citep{nakov2021clef,Shaar2021OverviewOT,barron2020overview,atanasova2018overview,konstantinovskiy2021toward,hassan2015detecting} filter claims from a source (similar to claim detection, sources include social media feeds and political debate transcripts, among others) by annotating claims based on the checkworthiness criteria (mentioned in the section  \ref{list:checkworthiness}). Each claim is given a checkworthiness score to obtain a ranked list. 
Note that claim detection and checkworthiness datasets may be expert annotated \citep{hassan2015detecting} or crowd annotated \citep{nakov2021clef,Shaar2021OverviewOT,barron2020overview,atanasova2018overview,konstantinovskiy2021toward}\footnote{\added{Some of these datasets, such as the CheckThat! datasets, are partially crowd and partially expert annotated.}}.

The datasets discussed above do not capture multi-modal datasets, and few do. One such dataset is r/Fakeddit \citep{Nakamura2020FakedditAN}. This dataset contains images and associated text content from Reddit as claims. Misinformation can also spread through multi-modal memes, and tasks such as Facebook (now Meta)  \textit{Hateful Memes Challenge} \citep{Kiela2020TheHM} for hate speech suggest what might be similarly done for misinformation detection.

\subsection{Evidence}


Early datasets in fact-checking provide metadata with claims as the only form of evidence. Such metadata include social media post properties, user information, publication date, source information \citep{Wang2017LiarLP,potthast2018stylometric}. As discussed earlier in Section \ref{subsec:task-evidence}, such metadata does not contain the world knowledge necessary to reason about a complex claim. To address the above limitations, recent datasets consider external evidence \citep{Guo2021ASO}. 

Evidence is collected differently depending upon the problem setup. For artificial claims, evidence is often retrieved from a single source such as Wikipedia articles \citep{thorne2018fever,jiang2020hover,schuster2021get}. Domain limited evidence for real-world claims is collected from problem-specific sources, such as academic articles for scientific claims \citep{Kotonya2020ExPubhealth,wadden2020fact}, or specific evidence listed in fact-checking websites \citep{vlachos2014fact,hanselowski2019richly}. Open-domain evidence for real-world claims is usually collected from the web via search engines  ~\citep{popat2018declare,Augenstein2019MultiFCAR}. 

Recently, there has been more work considering evidence beyond free text. Such formats include structured or semi-structured forms of evidence. Sources include knowledge bases for structured form of evidence \citep{shi2016discriminative} and semi-structured evidence from semi-structured knowledge bases \citep{vlachos-riedel-2015-identification}, tabular data \citep{chen2019seeing,gupta2020infotabs}, and tables within a document \citep{Aly2021FEVEROUSFE}. 

Additionally, there are some retrieval-specific datasets that aim at retrieving credible information from search engines \citep{TRECHeal37}. However, such tasks don’t incorporate claim checking as an explicit task.

\subsection{Veracity Prediction}

Evidence retrieval and veracity prediction datasets are usually constructed jointly. Note, in some cases, evidence may be absent from the datasets. Veracity prediction datasets usually do not deal with claim detection or claim checkworthiness tasks separately. Instead, such datasets contain a set of claims that are either artificially constructed or collected from the internet. 

Artificial claims in veracity prediction datasets are often limited in scope and constructed for natural language reasoning research \citep{Aly2021FEVEROUSFE,thorne2018fever,schuster2021get,jiang2020hover,chen2019tabfact}. For example, FEVER \citep{thorne2018fever} and HoVer \citep{jacovi2020aligning} obtain claims from Wikipedia pages. Some datasets also implement subject-predicate-object triplets for fact-checking against knowledge bases \citep{kim2020unsupervised,shi2016discriminative}.  

Fact-checking websites are popular sources for creating veracity prediction datasets based on real claims. Several datasets obtain claims from either a single website or collect claims from many such websites and collate them \citep{Wang2017LiarLP,Hanselowski2018ARA,Augenstein2019MultiFCAR,vlachos2014fact}. Note that such claims are inherently expert annotated. Other sources of claims are social media \citep{potthast2018stylometric,shu2017fake}, news outlets \citep{horne2018sampling,gruppi2021nela,norregaard2019nela}, blogs, discussions in QA forums, or similar user-generated publishing platforms \citep{mihaylova2018fact}. 

Additionally some fact-checking datasets target domain-specific problems such as scientific literature \citep{wadden2020fact}, climate change \citep{diggelmannclimate}, and public health \citep{Kotonya2020ExPubhealth}. Most datasets are \replaced{monolingual}{concerned with English language claims, however,} but recent effort have started to incorporate multi-lingual claims \citep{Gupta2021XFACTAN,barnabo2022fbmultilingmisinfo}. 

Early datasets focus on a binary veracity prediction - true or false \citep{mihalcea-strapparava-2009-lie}. Recent datasets often adopt an ordinal veracity labeling scheme that mimics fact checkering websites \citep{vlachos2014fact,Wang2017LiarLP,Augenstein2019MultiFCAR}. However, every fact-checking website has a different scale for veracity, so datasets that span across multiple websites come with a normalization problem. While some datasets do not normalize the labels \citep{Augenstein2019MultiFCAR}, some normalize them post-hoc \citep{Kotonya2020ExplainableAF,Gupta2021XFACTAN,hanselowski2019richly}.

\subsection{Explanation}
While an explanation is tied to veracity prediction, only a few datasets explicitly address the problem of explainable veracity prediction \citep{Atanasova2020GeneratingFC,Kotonya2020ExPubhealth,Alhindi2018WhereIY}. Broadly in NLP, often parts of the input is highlighted to provide an explanation for the prediction. This form of explanations is known as extractive rationale \citep{zaidan2007using,kutlu2020annotator}. Incorporating the idea of the extractive rationale, some datasets include a sentence from the evidence along with the label \citep{thorne2018fever,Hanselowski2018ARA,wadden2020fact,schuster2021get}. Although such datasets do not explicitly define evidence as a form of explanation in such cases,  the line between evidence retrieval and explanation blurs if the evidence is the explanation. \added{ However, explanations are different from evidence in a few ways. Particularly, explanations need to be concise for user consumption, while evidence can be a collection of documents or long documents. Explanations are user sensitive. Consequently, evidence alone as a form of explanation might have some inherent assumption about the user that might not be understandable for different groups of users (e.g., experts vs. non-experts).
} 

\subsection{Challenges}

\paragraph{Claims}
Checkworthiness datasets are highly imbalanced, i.e., the number of checkworthy claims are relatively low compared to non-checkworthy claims \citep{williams2020accenture}. Datasets are also not generalizable due to their limited domain-specific context \citep{Guo2021ASO}. 
Additionally, \added{while existing datasets cover various languages such as English, Arabic, Spanish, Bulgarian, and Dutch, they are primarily monolingual. }
\deleted{existing datasets are primarily in English with some Arabic tasks \citep{Shaar2021OverviewOT}.} 
Consequently, building multilingual checkworthiness predictors is still challenging.
\deleted{due to a lack of resources. }
\added{Much of the data in check-worthiness datasets is not originally intended to be used in classification. The criteria used by different organizations when selecting which claims to check is often subjective and may not generalize outside of the particular organization.}

Some annotation practices can result in artifacts in the dataset. For example, artificially constructed false claims, such as a negation-based false claim in FEVER, can lead to artifacts in models \citep{schuster2021get}. Models do not generalize well beyond the dataset 
because they might overfit to the annotation schema \citep{bansal2019beyond}. One way to identify such blind spots is by using adversarial datasets for fact-checking. Such a 
setting is incorporated in FEVER 2.0 \citep{thorne2019adversarial}.  

Datasets constructed for research may not always capture how fact-checkers work in practice. This leads to limitations in the algorithms built on them. For example, interviews with fact-checkers report that they tend to consider a combination of contents of the posts and associated virality metrics (indicating reach) during fact-checking \citep{ArnoldPh:online}. However, most fact-checking datasets do not include virality metrics. 

\paragraph{Evidence Retrieval} 
Some datasets have been constructed by using a claim verbatim as a query and taking the top search results as evidence. However, some queries are better than others for retrieving desired information. Consequently, greater care might be taken in crafting effective queries or otherwise improving evidence retrieval such that resulting datasets are more likely to contain quality evidence for veracity prediction. Otherwise, poor quality
evidence becomes a bottleneck for the \replaced{quality of the models trained at }{accuracy of} the later stages in the fact-checking pipeline \citep{singh2021case}.

\paragraph{Veracity Prediction} 
A key challenge in veracity prediction datasets is that the labels are not homogeneous across fact-checking websites and normalizing might introduce noise. 

\paragraph{Explanation}
Some datasets include entire fact-checking articles as evidence and their summaries as the form of explanation \citep{Atanasova2020GeneratingFC,Kotonya2020ExPubhealth}. 
In such cases, ``explanation'' components assume an already available fact-checking article. Instead, providing abstractive summaries and explaining the reasoning process over the evidence would be more valuable.

\paragraph{Data Generation}
Recent years have seen an increasing interest in the use of data generation and data augmentation for various NLP tasks \citep{Liu2022WANLIWA,hartvigsen-etal-2022-toxigen,dhole2021nl,kovatchev2021can}. The use of synthetic data has not been extensively explored in the context of fact-checking.

\section{Automating Fact-checking} \label{sec:approaches}

NLP research in automated fact-checking has primarily focused on building models for different automated fact-checking tasks utilizing existing datasets. In the following section, we highlight the broad modeling strategies employed in the literature, with more detailed discussion related to explainable methods for automated fact-checking. 

\subsection{General NLP Capabilities} \label{subsec:nlp}

\paragraph{Claim Detection and Checkworthiness}
While claim detection is usually implemented as a classification task only, claim checkworthiness is typically implemented both as ranking \citep{Nakov2021AutomatedFF} and classification task \citep{zeng2021automated}. As discussed earlier in the task formulation Section (\ref{subsec:task-claim}), the broad task of claim detection can be broken down into sub-tasks of identifying claims, filtering duplicate claims, and prioritizing claims based on their checkworthiness. Another instance of identifying claims is detecting rumors in social media streams.   

Some early works in rumor detection focus on feature engineering from available metadata the text itself \citep{enayet-el-beltagy-2017-niletmrg,aker2017simple,zhou2020fake}. More advanced methods for claim detection involve LSTM and other sequence models \citep{kochkina-etal-2017-turing}. Such models are better at capturing the context of the text \citep{zubiaga2016analysing}. Tree-LSTM \citep{ma-etal-2018-rumor} and Hierarchical attention networks \citep{guo2018rumor} capture the internal structure of the claim or the context in which the claim appears. Additionally, graph neural network approaches can capture the related social media activities along with the text \citep{monti2019fake}. 

Similarly, early works in claim-checkworthiness utilize support vector machines using textual features and rank the claims in terms of their priorities \citep{hassan2017toward}. 
For example, \citet{konstantinovskiy2021toward} build a classification model for checkworthiness by collapsing the labels to checkable vs.\ non-checkable claim. They build a logistic regression model that uses word embeddings along with syntax based features (parts of speech tags, and named entities). Neural methods such as LSTM performed well in earlier checkworthiness shared tasks \citep{elsayed2019overview}. Additionally, \citet{atanasova2019automatic} show that capturing context helps with the checkworthiness task as well. Models such as RoBERTa obtained higher performance in the later edition of the {\em \deleted{CLEF! }CheckThat!} shared task \citep{williams2020accenture,martinez2021nlp} for English language claims. Fine-tuning such models for claim detection tasks has become more prevalent for claim checkworthiness in other languages as well \citep{hasanain2020bigir,williams2020accenture}.

Filtering previously fact-checked claims is a relatively new task in this domain. \citet{Shaar2020ThatIA} propose an approach using BERT and BM-25 to match claims against fact-checking databases for matching claims with existing databases. Additionally, fine-tuning RoBERTa on various fact-checking datasets resulted in high performance for identifying duplicate claims \citep{bouziane2020team}. Furthermore, a combination of pretrained model Sentence-BERT and re-ranking with LambdaMART performed well for detecting previously fact-checked claims \citep{nakov2021clef}.

\paragraph{Evidence Retrieval and Veracity Prediction}

Evidence retrieval and veracity prediction in the pipeline can be modeled sequentially or jointly. Similar to claim detection and checkworthiness models, early works use stylistic features and metadata to arrive at veracity prediction without external evidence \citep{Wang2017LiarLP,rashkin2017truth}.
Models that include evidence retrieval often use commercial search APIs or some retrieval approach such as TF-IDF, and BM25 \citep{thorne2018fever}. Similar to question-answering models, some works adopt a two-step approach. First a simpler model (TF-IDF or BM-25) is used at scale and then a more complex model is used for re-ranking after the initial pruning 
\citep{thorne2018fever,nie-etal-2019-revealing,hanselowski2019richly}. Additionally, document vs.\ passage retrieval, or 2-stage ``telescoping'' approaches, are adopted where the first stage is retrieving related documents and the second stage is to retrieve the relevant passage. Two stage approaches are useful for scaling up applications as the first stage is more efficient than the second stage. For domain specific evidence retrieval, using domain-bound word embeddings has been shown to be effective \citep{zeng2021automated}. 

The IR task is not always a part of the process. 
Instead, it is often assumed that reliable evidence is already available. While this simplifies the fact-checking task so that researchers can focus on veracity prediction, in practice evidence retrieval is necessary and cannot be ignored. Moreover, in practice one must contend with noisy (non-relevant), low quality, and biased search results during inference.

As discussed earlier in Section \ref{subsec:task-veracity}, assessing the reliability of gathered evidence may be necessary. If the evidence is assumed to be trustworthy, then it suffices to detect the stance of each piece of evidence and then aggregate (somehow) to induce veracity (e.g., perhaps assuming all evidence is equally important and trustworthy). However, often one must contend with evidence ``in the wild'' of questionable reliability, in which case assessing the quality (and bias) of evidence is an important precursor to using it in veracity prediction.



Veracity prediction utilizes textual entailment for inferring veracity over either a single document as evidence or over multiple documents. \citet{dagan2010recognizing} define {\em textual entailment} as ``deciding, given two text fragments, whether the meaning of one text is entailed (can be inferred) from another text.'' Real-world applications often require reasoning over multiple documents \citep{Augenstein2019MultiFCAR,Kotonya2020ExPubhealth,schuster2021get}. Reasoning over multiple documents can be done either by concatenation \citep{nie-etal-2019-revealing} or weighted aggregation \citep{Nguyen18-aaai}. Weighted aggregation virtually re-ranks the evidence considered to filter out the unreliable evidence \citep{ma-etal-2019-sentence,pradeep2020scientific}. Some approaches also use Knowledge Bases as the central repository of all evidence \citep{shi2016discriminative}. However, evidence is only limited to what is available in the knowledge base \citep{Guo2021ASO,zeng2021automated}. Moreover, a fundamental limitation of knowledge bases is that not all knowledge fits easily into structured relations.

Recent developments in large language models help extend the knowledge base approach. Fact-checking models can rely on pretrained models to provide evidence for veracity prediction \citep{lee-etal-2020-language}. However, this approach can encode biases present in the language model \citep{lee-etal-2021-towards}. 

An alternative approach is to help fact-checkers with downstream tasks by processing evidence. An example of such work is generating summaries over available evidence using BERT 
\citep{fan2020generating}. 

\paragraph{Limitations} 
With the recent development of large, pre-trained language models and deep learning for NLP, we see a significant improvement across the fact-checking pipeline. 
With the introduction of FEVER \citep{thorne2018fever,thorne2019fever2,Aly2021FEVEROUSFE} and 
{\em\deleted{ CLEF! }CheckThat!} \citep{nakov2021clef} we have benchmarks for both artificial and real-life claim detection and verification models. However, even the state-of-the-art NLP models perform poorly on the benchmarks above. For example, the best performing model on FEVER 2018 shared task \citep{thorne2018fever} reports an accuracy of 0.67\footnote{\url{https://fever.ai/2018/task.html}}. Models perform worse on multi-modal shared task FEVEROUS \citep{Aly2021FEVEROUSFE}: the best performing model reports 0.56 accuracy score\footnote{\url{https://fever.ai/task.html}}. Similarly, the best checkworthiness model only achieved an average precision of 0.65 for Arabic claims and 0.224 for English claims in the {\em\deleted{ CLEF! }CheckThat!} 2021 shared task for identifying checkworthiness in tweets \citep{nakov2021clef}. On the other hand, the best performing model for identifying check-worthy claims in debates reports 0.42 average precision. Surprisingly, \citet{barron2020overview}, the top performing model for checkworthiness detection, 
report an average precision of 0.806 \citep{williams2020accenture}. 
\deleted{Both  CLEF!\ CheckThat! 2020 and 2021 use the same dataset but new test sets for evaluation \citep{nakov2021clef,barron2020overview} }. 
For the fact-checking task \added{of CheckThat!\ 2021 \citep{nakov2021clef}}, the best performing model reports a 0.83 macro F1 score. However, the second-best model only reports a 0.50 F1 score. \replaced{Given this striking gap in performance between the top system vs.\ others, it would be valuable for future work to benchmark systems on additional datasets in order to better assess the generality of these findings}{It is unclear whether the best model is generalizable or overfits the particular dataset}. 

It is not easy 
to make a direct comparison between different methods that are evaluated in different settings and with different datasets \citep{zeng2021automated}. Moreover, the pipeline design of automated fact-checking creates potential bottlenecks, e.g., performance on the veracity prediction task on most datasets is dependent on the claim detection task performance or the quality of the evidence retrieved. Extensive benchmarks are required to incorporate all of the prior subtasks in the fact-checking pipeline systematically \citep{zeng2021automated}.

Much of AI research is faced with a fundamental trade-off between working with diverse formulations of a problem and standardized benchmarks for measuring progress. This trade-off also impacts automated fact-checking research. While there exist benchmarks such as FEVER and the {\em\deleted{ CLEF! }CheckThat!\deleted{ Lab}}, most models built on those benchmarks may not generalize well in a practical setting. Abstract and tractable formulations of a problem may help us develop technologies that facilitate practical adoption. However, practical adoption requires significant engineering effort beyond the research setting. Ideally, we would like to see automated fact-checking research continue to move toward increasingly realistic benchmarks while incorporating diverse formulations of the problem.  




\subsection{Explainable Approaches} \label{subsec:explainable}


Although the terms \textit{interpretability} and \textit{explainability} are often used interchangeably, and some times defined to be so \citep{molnar2020interpretable},
we distinguish interpretability vs.\ explainability similar to \citep{Kotonya2020ExplainableAF}. Specifically, {\em interpretability} represents methods that provide direct insight into an AI system’s components (such as features and variables), often requiring some understanding of the specific to the algorithm, and often built for expert use cases such as model debugging. {\em Explainability} represents methods to understand an AI model without referring to the actual component of the systems. Note that, in the task formulation section, we have also talked about explaining veracity prediction. The goal of such explanation stems from fact-checker needs to help readers understand the fact-checking verdict. Therefore, explaining veracity prediction aligns more closely with explainability over interpretability. When the distinction between explainability vs.\ interpretability does not matter, we follow \citet{vaughan2020human} in adopting 
{\em intelligibility} \citep{vaughan2020human} as an umbrella term for both concepts. 

\citet{sokol2019desiderata} propose a desiderata for designing user experience for machine learning applications. \citet{Kotonya2020ExplainableAF} extend them in the context of fact-checking and suggest eight properties of intelligibility: {\em actionable}, {\em causal}, {\em coherent}, {\em context-full}, {\em interactive}, {\em unbiased or impartial}, {\em parsimonious}, and {\em chronological}.

Additionally, there are three dimensions specifically for explainable methods in NLP \citep{jacovi-goldberg-2020-towards}:
\begin{enumerate}
    \item \textbf{Readability:} are explanations clear?
    \item \textbf{Plausibility:} are explanations compelling or persuasive?
    \item \textbf{Faithfulness:} are explanations faithful to the model's actual reasoning process?
\end{enumerate}

In comparison with the available intelligibility methods in NLP \citep{Wiegreffe2021TeachMT}, only a few are applied to existing fact-checking works. Below, we highlight only commonly observed explainable fact-checking methods (also noted by \citet{Kotonya2020ExplainableAF}). 

\paragraph{Attention-based Intelligibility} Despite the debate about attention being a reliable intelligibility method \citep{jain2019attention,wiegreffe2019attention,serrano2019attention,bibal-etal-2022-attention}, it remains a popular method in existing deep neural network approaches in fact-checking. Attention-based explanations are provided in various forms:
\begin{enumerate}
    \item highlighting tokens in articles \citep{popat2018declare}
    \item highlighting salient excerpts from evidence utilizing comments related to the post \citep{shu2019defend}
    \item n-gram extraction using self-attention \citep{yang2019xfake}
    \item attention from different sources other than the claim text itself, such as the source of tweets, retweet propagation, and retweeter properties \citep{lu2020gcan}
\end{enumerate}
\paragraph{Rule discovery as explanations} Rule mining is a form of explanation prevalent in knowledge base systems \citep{gad2019exfakt,ahmadi2019explainable}. These explanations can be more comprehensive, but as noted in the previous section, not all statements can be fact-checked via knowledge-based methods due to limitations of the underlying knowledge-base itself. Some approaches provide general purpose rule mining in an attempt to address this limitation \citep{ahmadi2020rulehub}.

\paragraph{Summarization as explanations} Both extractive and abstractive summaries can provide explanations for fact-checking. \citet{Atanasova2020GeneratingFC} provides natural language summaries to explain the fact-checking decision. They explore two different approaches - explanation generation and veracity prediction as separate tasks, and joint training of the both. Joint training performs worse than single training. \citet{Kotonya2020ExPubhealth} combine abstractive and extractive approaches to provide a novel summarization approach. \added{\cite{brand2018neural} show jointly training prediction and explanation generation with encoder-decoder models such as BART \citep{lewis2020bart} results in explanations that help the crowd to perform better veracity assessment.}

\paragraph{Counterfactuals and adversarial methods} Adversarial attacks on opaque models help to identify any blind-spots, biases and discover data artifacts in models \citep{ribeiro2020beyond}. Shared task FEVER 2.0 \citep{thorne2019fever2} asked participants to devise methods for generating adversarial claims to identify weaknesses in the fact-checking methods. Natural language generation models such as GPT-3 can assist in formulating adversarial claims. More control over the generation can come from manipulating the input to natural language generation methods and constraining the generated text within original vocabulary \citep{niewinski2019gem}. \citet{atanasova2020generating} generate claims with n-grams inserted into the input text. \citet{thorne2019adversarial} experiment with several adversarial methods such as rule-based adversary, semantically equivalent adversarial rules (or SEARS) \citep{ribeiro2018semantically}, negation, and paraphrasing-based adversary. Adversarial attacks are evaluated based on the {\em potency} (correctness) of the example and reduction in system performance. While methods such as SEARS and paraphrasing hurt the system performance, hand-crafted adversarial examples have higher potency score.

\paragraph{Interpretable methods (non-BlackBox)} Some fact-checking works use a white-box or inherently interpretable model for fact-checking. \citet{Nguyen18-aaai,nguyen2018believe} utilize a probabilistic graphical model and build an interactive interpretable model for fact-checking where users are allowed to directly override model decisions. \citet{kotonya2021graph} propose an interpretable graph neural network for interpretable fact-checking on FEVEROUS dataset \citep{Aly2021FEVEROUSFE}.

\paragraph{Limitations} Intelligible methods in NLP and specifically within fact-checking are still in their infancy. 
Analysis of \citet{Kotonya2020ExplainableAF} shows that most methods do not fulfill the desiderata mentioned earlier in this section. Specifically, they find that none of the existing models meet 
the criteria of being actionable, causal, and chronological. They also highlight that no existing method explicitly analyzes whether explanations are impartial. Some forms of explanations, such as rule-based triplets, are unbiased as they do not contain sentences or contain fragments of information \citep{Kotonya2020ExplainableAF}. 

Some explainable methods address a specific simplified formulation of the task. For example, \citet{Kotonya2020ExPubhealth, Atanasova2020GeneratingFC} both assume that expert-written fact-checking articles already exist. They provide explanations as summaries of the fact-checking article. However, in practice, a fact-checking system would not have access to such an article for an unknown claim. 

In the case of automated fact-checking, most intelligible methods focus on explaining the outcome rather than describing the process to arrive at the outcome \citep{Kotonya2020ExplainableAF}. Moreover, 
all of the tasks in the fact-checking pipeline have not received equal attention for explainable methods. \citet{Kotonya2020ExplainableAF} also argue that automatic fact-checking may benefit from explainable methods that provide insight into how outcomes of earlier sub-task in the fact-checking pipeline impact the outcome of later subtasks. 

Most explainable NLP works evaluate explanation quality instead of explanation utility or faithfulness. \citet{jacovi-goldberg-2020-towards} argue for a thorough faithfulness evaluation for explainable models. For example, even though attention-based explanations may provide quality explanations, they may not necessarily be faithful. Moreover, explanation utility requires separate evaluation by measuring whether explanations improve both i) human understanding of the model \citep{Hase2020EvaluatingEA} and ii) human effectiveness of the downstream task \citep{Nakov2021AutomatedFF}. Additionally, most intelligible methods establish only one-way communication from the model to humans. Instead, explanations might improve the model and human performance by establishing a bidirectional feedback loop. 


\subsection{Human-in-the-loop Approaches} \label{subsec:hitl}

Human-in-the-loop (HITL) approaches can help scale automated solutions while utilizing human intelligence for complex tasks. There are different ways of applying HITL methods, e.g., delegating sub-tasks to crowd workers \citep{demartini2013crowdq,demartini2012zencrowd,sarasua2012crowdmap}, active learning \citep{settles2009active,zhang2017active}, interactive machine learning \citep{amershi2014power,joachims2007search}, and decision support systems where humans make the final decision based on model outcome and explanations \citep{zanzotto2019human}. 

While HITL approaches in artificial intelligence are prevalent, only a few recent works employ such approaches in fact-checking. HITL approaches are predominantly more present in the veracity prediction task than other parts of the pipeline. For example, \citet{demartini2020human} propose a HITL framework for combating online misinformation. However, they only consider hybrid approaches for two sub-tasks in the fact-checking pipeline: a) claim check-worthiness and b) truthfulness judgment (same as veracity prediction). Below, we discuss the existing HITL approaches by how the system leverages human effort for each sub-task in the fact-checking pipeline.


\paragraph{Claim Detection, Checkworthiness, and Prioritization} Social media streams are often monitored for rumors as a part of the claim detection task \citep{Guo2021ASO}. \citet{farinneya-etal-2021-active-learning} apply an active learning-based approach at the claim detection stage for identifying rumors on social media. In-domain data is crucial for traditional supervised methods to perform well for rumor detection \citep{ahsan2019detection}, but in real-world scenarios, sufficient in-domain labeled data may not be available in the early stages of development. A semi-supervised approach such as active learning is beneficial for achieving high performance with fewer data points. Empirical results shows that Tweet-BERT, along with the least confidence-based sample selection approach, performs on par with supervised approaches using far less labeled data \citep{farinneya-etal-2021-active-learning}. 


Similarly, \citet{tschiatschek2018fake} propose a HITL approach that aims to automatically aggregate user flags and recommend a small subset of the flagged content for expert fact-checking. 
Their Bayesian inference-based approach jointly learns to detect fake news and identify the accuracy of user flags over time. One strength of this approach is that the algorithm improves over time in identifying users' flagging accuracy. Consequently, over time this algorithm's performance improves. This approach is also robust against spammers. By running the model on publicly available Facebook data where a majority of the users are adversarial, experiments show that their algorithm still performs well. 


Duke's Tech \& Check team implemented HITL at the claim check-worthiness layer \citep{Alessoni87:online}. To avoid flagging false check-worthy claims, a human expert would sort claims detected by ClaimBuster \citep{hassan2017claimbuster}, filter out the ones deemed more important for fact-checkers, and email them to several organizations. In essence, this approach helped fact-checkers prioritize the claims to check through an additional level of filtering. Currently, several published fact-checks on PolitiFact were first alerted by the emails from Tech \& Check.

Note that the {\em\deleted{ CLEF! }CheckThat!\deleted{ Lab}} \citep{nakov2021clef,Shaar2021OverviewOT,barron2020overview,atanasova2019overview} is a popular shared task for claim detection, check-worthiness, and prioritization tasks. However such shared tasks often have no submissions that employs HITL methodologies. Shared tasks for HITL approaches could encourage more solutions that can complement the limitations of model-only based approaches.

\paragraph{Evidence Retrieval and Veracity Prediction} Most work in HITL fact-checking caters to veracity prediction, and only a few consider evidence retrieval as a separate task. While there is a body of literature on HITL approaches in information retrieval \citep{chen2013improving,demartini2015hybrid}, we know of no work in that direction for fact-checking. 

\citet{shabani2021sams} leverage HITL approaches for providing feedback about claim source, author, message, and spelling (SAMS). Annotators answer four yes/no questions about whether the article has a source, an author, a clear and unbiased message, and any spelling mistake. Furthermore, this work integrates the features provided by humans in a machine learning pipeline, which resulted in a 7.1\% accuracy increase. However, the evaluation is performed on a small dataset with claims related to Covid-19. It is unclear if this approach would generalize outside of the domain. Moreover, further human effort can be reduced in this work by automating spell-check and grammar-check. SAMS could be quite limited in real life situations as most carefully crafted misinformation often looks like real news. Model generated fake news can successfully fool annotators \citep{zellers2020defending}, and thus SAMS might also fail to flag such fake news. 


\replaced{
\citet{qu2021combining} and \citet{quhuman}}
{\citet{quhuman}} 
provide an understanding of how human and machine confidence scores can be leveraged to build HITL approaches for fact-checking. They consider explicit self-reported annotator confidence and compute implicit confidence based on standard deviation among ten crowd workers. Model confidence is obtained from bootstrapping \citep{efron1985bootstrap} ten different versions of the model and then computing standard deviation over the scores returned by the soft-max layer. Their evaluation shows that explicit crowd and model confidence are poor indicators of accurate classification decisions. Although the crowd and the model make different mistakes, there is no clear signal that confidence is related to accuracy. However, they show that implicit crowd confidence can be a useful signal for identifying when to engage experts to collect labels.
A more recent study shows that a politically balanced crowd of ten is correlated with the average rating of three fact-checkers \citep{allen2020scaling}. \citet{gold2019annotating} also find that annotations by a crowd of ten correlate with the judgments of three annotators for textual entailment, which is utilized by veracity prediction models.

A series of studies show that the crowd workers can reliably identify misinformation \citep{roitero2020can,roitero2020covid,soprano2021many}. Furthermore, \citet{roitero2020covid} show that crowd workers not only can identify false claims but also can retrieve proper evidence to justify their annotation. One weakness of this study is that it only asks users to provide one URL as evidence. However, in practice, fact-checking might need reasoning over multiple sources of information. Although these studies do not propose novel HITL solutions, they provide sufficient empirical evidence and insights about where crowd workers can be engaged reliably in the fact-checking pipeline. 


\citet{Nguyen18-aaai} propose joint modeling of crowd annotations and machine learning to detect the veracity of textual claims. The key strength of the model is that it assumes all annotators can make mistakes, which is a possibility as fact-checking is a difficult task. Another strength is that this model allows users to import their knowledge into the system. Moreover, this HITL approach can collect on-demand stance labels from the crowd and incorporate them in veracity prediction. Empirical evaluation shows that this approach achieves strong predictive performance. A follow-up study provides an interactive HITL tool for fact-checking \citep{nguyen2018believe}. 


\citet{nguyen2020factcatch} propose a HITL system to minimise user effort and cost. Users validate algorithmic predictions but do so at a minimal cost by only validating the most-beneficial predictions for improving the system. This system provides a guided interaction to the users and incrementally gets better as users engage with it. 


It is important to note that research on crowdsourcing veracity judgment is at an early stage. Different factors such as demographics, political leaning, criteria for determining the expertise of the assessors \citep{bhuiyan2020investigating}, cognitive factors \citep{kaufman2022s}, and even the rating scale \citep{la2020crowdsourcing} led to different levels of alignment with expert ratings. \citet{bhuiyan2020investigating} outline research directions for designing better crowd processes specific to different types of misinformation for the successful utilization of crowd workers. 

\paragraph{Explaining Veracity Prediction} HITL systems in fact-checking often use veracity explanations to correct model errors. As discussed earlier, \citet{nguyen2018believe} provides an interpretable model that allows users to impart their knowledge when the model is wrong. Empirical evaluation shows that users could impart their knowledge into the system. Similarly, \citet{zhang2021faxplainac} propose a method that collects user feedback from explanations. Note that this method explains veracity prediction outcomes based on the evidence retrieved and their stance. Users provide feedback in terms of stance and relevance of the retrieved evidence. The proposed approach employs lifelong learning which enables the system to improve over time. Currently there is no empirical evaluation of this system to identify the effectiveness of this approach. 

Although natural language generation models are getting increasingly better \citep{radford2019language}, generating abstractive fact-checking explanations is still in its infancy \citep{Kotonya2020ExPubhealth}. HITL methods could be leveraged to write reports justifying fact-checking explanations.



\paragraph{Limitations} After reviewing existing HITL approaches across different fact-checking tasks, we also list out several limitations as follow. First, some HITL approaches adopt several interpretable models to integrate human input, but the resulting models do not perform as well as the state-of-the-art deep learning models \citep{Nguyen18-aaai, nguyen2018believe}. 

\citet{farinneya-etal-2021-active-learning} apply HITL approaches to scale up rumor detection from a limited amount of annotated data. Although it performs well to generalize the algorithm for a new topic in a few-shot manner, one of the weaknesses is that data from other domains or topics causes a high variance in model performance. Consequently, in-domain model performance might degrade when out-of-domain data is introduced in model training. This issue may hinder the model's generalizability in practice, especially where a clear demarcation between topic domains may not be possible.

More importantly, there is a lack of empirical studies on how to apply HITL approaches of fact-checking for practical adoption. Although HITL approaches provide a mechanism to engage human in the process of modeling development, several human factors, such as usability, intelligibility, and trust, become important to consider when applying this method in the real-world use case. Fact-checking is a time-sensitive task and requires expertise to process complex information over multiple sources \citep{graves2017anatomy}. Fact-checkers and policy makers are often skeptical about any automated or semi-autoamted solutions as this type of research requires human creativity and expertise \citep{ArnoldPh:online, Micallef2022}. Therefore, more empirical evidence needs to be found to assess the effectiveness of applying different HITL approaches to automated fact-checking.

\section{Existing Tools for Fact-checking} \label{sec:tools}

In the previous section, we reviewed the details of current NLP technologies for fact-checking. Subsequently, we extend our review of automated fact-checking to the HCI literature and discuss existing practices of applying fact-checking into real-world tools that assist human fact-checkers. In brief, there is a lack of holistic review of fact-checking tools from a human-centered perspective. Additionally, we found that the articulation of work between human labor and AI tools is still opaque in this field. Research questions include but are not limited to: 1) how can NLP tools facilitate human work in different fact-checking tasks? 2) how can we incorporate user needs and leverage human expertise to inform the design of automated fact-checking?  

In this section, we examine current real-world tools that apply NLP technologies in different stages of fact-checking and clarify the main use cases of these tools. We argue that more research concerning human factors for building automated fact-checking, such as user research, human-centered design, and usability studies, should be conducted to improve the practical adoption of automated fact-checking. These studies help us identify the design space of applying explainable and HITL approaches for real-world NLP technologies.

\subsection{Claim Detection and Prioritization} \label{subsec:decision-making}
\added{The first step in claim detection is sourcing content to possibly check.  On end-to-end encrypted platforms, such as WhatsApp, Telegram, and Signal, crowdsourcing-based tip-lines play a vital role in identifying suspicious content that is not otherwise accessible \citep{Kazemi2021TiplinesTC}. As another example, {\em Check} from Meedan}
\footnote{\url{https://meedan.com/check}}, 
\added{a tip-line service tool, also helps fact-checkers monitor fake news for in-house social media. User flagging of suspect content on social media platforms such as Facebook is also a valuable signal for identifying such content, and crowdsourcing initiatives like Twitter's BirdWatch can further help triage and prioritize claims for further investigation.} 

In the stage of finding and choosing claims to check, fact-checkers assess the fact-checking related quality of a claim and decide whether to fact-check it \citep{graves2017anatomy, Micallef2022}. NLP models in claim detection, claim matching, and check-worthiness are useful to assist the above decision-making process. However, integrating them into real-world tools that help fact-checkers prioritize what to check requires more personalized effort. \citet{graves2018understanding} points out that it is important to design the aforementioned models to cater to fact-checker organizational interests, stakeholder needs, and changing news trends.

As one of the fact-checking qualities of a claim, checkability can be objectively analyzed by whether a claim contains one or more purported facts that can be verified (Section \ref{subsec:task-claim}). Fact-checkers find it useful to apply models that identify checkable claims to their existing workflow because the model helps them filter irrelevant content and claims that are uncheckable when they are choosing claims to check \citep{ArnoldPh:online}. ClaimBuster, one of the well-known claim detection tools, is built to find checkable claims from a large scale of text content \citep{hassan2017toward}. Claim detection can also be integrated into speech recognition tools to spot claims from live speech \citep{Adair2020}. 

Additionally, if a claim has already been fact-checked, fact-checkers can skip it and prioritize claims that have not been checked. As a relatively new NLP task, claim matching has been integrated into some current off-the-shelf search engines or fact-checking tools to help fact-checkers find previously fact-checked claims. For example, Google Fact Check Explorer\footnote{\url{https://toolbox.google.com/factcheck/explorer}} can retrieve previously fact-checked claims by matching similar fact-check content to user input queries. Similarly, with Meedan's {\em Check}, if users send a tip with fake news that has been previously fact-checked, the tool further helps fact-checkers retrieve the previous fact-check and send it to users.  

\added{Whether or not to fact-check a claim depends on an organization's goals and interests.} \deleted{However, to identify whether or not to fact-check a claim such as check-worthiness – journalists refer to it as newsworthiness, designing tools with corresponding NLP task should cater to fact-checkers’ organizational interest and requires more human involvements to tailor the system.} \added{Tools built for claim detection need to take such interests into account.} For example, Full Fact developed a \deleted{sophisticated} claim detection system that classifies claims into different categories, such as quantity, predictions, correlation or causation, personal experience, and laws or rules of operations \citep{konstantinovskiy2021toward}. The claim categories are designed by their fact-checkers to cater to their needs of fact-checking UK political news in a live fact-checking situation. Identifying certain claims, such as quantity, correlation or causation, might be particularly useful for fact-checkers to evaluate the credibility of politician statements and claims. The system also helps tailor fact-checkers' downstream tasks, such as fact-check assignments and automated verification for statistical claims \citep{Nakov2021AutomatedFF}.

Fact-checkers also use social media monitoring tools to find claims to check, such as CrowdTangle, TweetDeck, and Facebook's (unnamed) fact-checking tool, but those tools are not very effective to detect checkable claims. Some fact-checkers reported that only roughly 30\% of claims flagged by Facebook's fact-checking tool were actually checkable \citep{ArnoldPh:online}. A low hanging fruit is to integrate claim detection models into these social media monitoring tools so that it is easier for fact-checkers to identify claims that are both viral and checkable. Additionally, these tools should enable fact-checkers to locate certain figures, institutions, or agencies according to their fact-checking interests and stakeholder needs so that these tools can better identify and prioritize truly check-worthy claims.
\added{An important question in implementing those systems is how to measure the virality of a claim and its change over time.}

It would also be useful to integrate veracity prediction into previous fact-checking tools because fact-checkers may pay the most attention to claims\footnote{\url{https://www.factcheck.org/our-process/}} that are suspect and uncertain (since obviously true or false claims likely do not require a fact-check). However, information or data points that are used to give such predictions should also be provided to fact-checkers. If sources, evidence, propagation patterns, or other contextual information that models use to predict claim veracity can be explained clearly for fact-checkers, they can also triage these indicators to prioritize claims more holistically.

\subsection{Tools for Evidence Retrieval} \label{subsec:information-seeking}

After finding and prioritizing which claim to check, fact-checkers investigate claims following three main activities: 1) decomposing claims, 2) finding evidence, and 3) tracing the provenance of claims and their spread. Note that these three activities are intertwined with each other by using different information-seeking tools in the fact-checking process. Fact-checkers search for evidence by decomposing claims into sub questions. Evidence found while investigating a claim may further modify or add to the sub-questions \citep{singh2021case}. By iteratively investigating claims via online search, fact-checkers reconstruct the formation and the spread of a claim to assess its truth \citep{graves2017anatomy}. In this section, we discuss the utility of existing information-seeking tools, including off-the-shelf search engines and domain-specific databases, that assist fact-checkers in each activity.

Claim decomposition is not a specific activity that qualitative researchers have reported or analyzed in their fact-checking studies, but we can find more details from where fact-checking organizations describe their methodology\footnote{\url{https://leadstories.com/how-we-work.html}} and how fact-checkers approach complex claims in their fact-checks\footnote{\url{https://www.factcheck.org/2021/10/oecd-data-conflict-with-bidens-educational-attainment-claim/} In this fact-check, fact-checkers decompose what President Biden mean by “advanced economies” and “young people”. The approach of defining these two terms directly influence their fact-checking results.}. Claim decomposition refers to how fact-checkers interpret ambiguous terms of a claim and set the fact-checking boundaries to find evidence. Decomposing claims effectively requires sensitive curiosity and news judgments for fact-checkers that are cultivated through years of practice. Unfortunately, we are not aware of any existing tools that facilitate this process. 

Traditional methodology to decompose claims is to ask sub-questions. Recent NLP studies simulate this process by formulating it as a question-answering task \citep{fan2020generating, chen2022generating}. Researchers extract justifications from existing fact-checks and crowdsource sub-questions to decompose the claim. For automated-fact-checking, this NLP task might be very beneficial to improve the performance of evidence retrieval by auto-decomposing claims into smaller checkable queries \citep{chen2022generating}. Although it is difficult for NLP to match the abilities of professional fact-checkers, it might help scale up the traditional, human fact-checking process. It could also help the public, new fact-checkers, or journalists to more effectively investigate complex claims and search for evidence.

How fact-checkers find evidence is usually a domain-specific reporting process, contacting experts or looking for specific documents from reliable sources \citep{graves2017anatomy, Micallef2022}. Instead of conducting random searches online, most fact-checkers include a list of reliable sources in which to look for evidence. Tools that are designed for searching domain datasets can also help fact-checkers to find evidence. For example, \citet{Li2021} built an analytical search engine for retrieving the COVID-19 news data and summarizing it in an easy to digest, tabular format. The system can decompose analytical queries into structured entities and extract quantitative facts from news data. Furthermore, if evidence retrieval is accurate enough for in-domain datasets, the system can take a leap further to auto-verify domain-related claims. We provide more detailed use cases of veracity prediction in Section \ref{subsec:domain}.

Fact-checkers mainly use off-the-shelf search engines, such as Google, Bing, etc., to trace a claim's origin from publicly accessible documents \citep{Beers2020, ArnoldPh:online}. Other digital datasets, such as \textit{LexisNexis} and \textit{InternetArchive}, are also useful for fact-checkers to trace claim origin. To capture the formation and change of a claim, search engines should not only filter unrelated content, but also retrieve both topically and evidentially relevant content. \citet{Hasanain2021} report that most topically relevant pages retrieved from Google do not contain evidential information, such as statistics, quotes, entities, or other types of facts. Additionally, most built-in search engines in social media platforms, such as Twitter and Facebook, only filter “spreadable'' content not “credible” content \citep{Alsmadi2021}.

Furthermore, these off-the-shelf search engines do not support multilingual search, so it is difficult for fact-checkers to trace claims if they are translated from other languages \citep{graves2017anatomy, Nakov2021AutomatedFF}. NLP researchers have started to use multilingual embedding models to represent claim-related text in different languages and match existing fact-checks \citep{Kazemi2021multilingual}. This work not only helps fact-checkers find previously fact-checked claims more easily from other languages, but also to examine how the claim is transformed and reshaped by the media in different languages and socio-political contexts.

\subsection{Domain-specific Tools for Claim Verification} \label{subsec:domain}

As discussed in Sections \ref{subsec:task-evidence} and \ref{subsec:task-veracity}, most verification prediction models are grounded on the collected evidence and the claim. To build an end-to-end claim verification system, NLP developers need to construct domain-specific datasets incorporating both claims and evidence. Different from complex claims that contain multiple arguments and require decomposition, claims that have simple linguistic structure with purported evidence or contain statistical facts can be automatically verified \citep{Nakov2021AutomatedFF}. 

\citet{karagiannis2020scrutinizer} built CoronaCheck, a search engine that can directly verify Covid-19 related statistical claims by retrieving official data curated by experts \citep{dong2020interactive}. Full Fact \citep{ThePoynterInstitute2021} also took a similar approach to verify statistical macroeconomic claims by retrieving evidence from UK parliamentary reports and national statistics. Additionally, \citet{wadden2020fact} built a scientific claim verification pipeline by using abstracts that contain evidence to verify a given scientific claim.

However, pitfalls still exist if fact-checkers use these domain-specific verification tools in practice. For example, the CoronaCheck tool cannot check the claim “The Delta variant causes more death than the Alpha variant” simply because the database does not contain fine-grained death statistics for Covid variants. Additionally, checking a statistical or scientific claim might only be a part of the process of checking a more complex claim, which requires fact-checkers to contextualize the veracity of previous statistical or scientific checks. In general, domain-specific tools are clearly valuable to use when available, though in practice they are often incomplete and insufficient on their own to check complex claims. 

\section{Discussion} \label{sec:future}

In this review, we have 1) horizontally outlined the research of applying NLP technologies for fact-checking from the beginning of task formulation to the end of tools adoption; as well as 2) vertically discussing the capabilities and limitations of NLP for each step of a fact-checking pipeline. We perceive a lack of research that bridges both to assist fact-checkers. Explainable and HITL approaches leverage both human and computational intelligence from a human-centered perspective, but there is a need to provide actionable guides to utilize both methods for designing useful fact-checking tools. In this section, we propose several research directions to explore the design space of applying NLP technologies to assist fact-checkers.

\subsection{Distributing Work between Human and AI for Mixed-initiative Fact-checking} \label{subsec:articulation}

The practice of fact-checking has already become a type of complex and distributed computer-mediated work \citep{graves2018understanding}. Although \citet{graves2017anatomy} breaks down a traditional journalist fact-checking pipeline into five steps, the real situation of fact-checking a claim is more complicated \citep{juneja2022human}. Various AI tools are adopted dynamically and diversely by fact-checkers to complete different fact-checking tasks \citep{ArnoldPh:online, Beers2020, Micallef2022}. 

Researchers and practitioners increasingly believe that future fact-checking should be a mixed-initiative practice in which humans perform specific tasks while machines take over others \citep{nguyen2018believe, lease2020designing, Nakov2021AutomatedFF}. To embed such hybrid and dynamic human-machine collaborations into existing fact-checking workflow, the task arrangement between human and AI need to be articulated clearly by understanding the expected outcomes and criteria for each. Furthermore, designing a mixed-initiative tool for different fact-checking tasks requires a more fine-grained level of task definition for human and AI \citep{lease2018fact, lease2020designing}. In Section \ref{subsec:hitl}, we discuss several studies highlighting the role of humans in the fact-checking workflow, e.g.,  a) human experts select check-worthy claims from claim detection tools \citep{hassan2017claimbuster} and deliver them to fact-checkers, b) ask crowd workers to judge reliable claims sources \citep{shabani2021sams}, or c) flag potential misinformation \citep{roitero2020covid} to improve veracity prediction. All of the above human activities are examples of micro-tasks within a mixed-initiative fact-checking process.        

Prior work in crowdsourcing has shown that it is possible to effectively break down the academic research process and utilize crowd workers to partake in smaller research tasks \citep{vaish2015crowdsourcing,vaish2017crowd}. Given this evidence, we can also break down sub-tasks of a traditional fact-checking process into more fine-grained tasks. Therefore, key research questions include: a) How can we design these micro-tasks to facilitate each sub-task of fact-checking, and b) What are the appropriate roles for human and AI in different micro-tasks? 

To effectively orchestrate human and AI work, researchers need to understand the respective roles of human and AI, and how they will interact with one another, because it will directly affect whether humans decide to take AI advice \citep{Cimolino2022sharedcontrol}.
Usually, if AI aims to assist high-stake decision-making tasks, such as recidivism prediction \citep{veale2018fairness} and medical treatments \citep{cai2019hello}, considerations of risk and trust will be important factors for people to adopt such AI assistants \citep{lai2021towards}. In the context of fact-checking, if AI directly predicts the verdict of a claim, fact-checkers may be naturally skeptical about how the AI makes such a prediction \citep{ArnoldPh:online}. On the other hand, if AI only helps to filter claims that are uncheckable, such as opinions and personal experience, fact-checkers may be more willing to use such automation with less concern about how AI achieves it. Deciding whether a claim is true or false is a high-stake decision-making task for fact-checkers, while filtering uncheckable claims is a less important but tedious task that fact-checkers want automation to help with. Therefore, the extent of human acceptance of AI varies according to how humans assess the task assigned to AI, resulting in different human factors, such as trust, transparency, and fairness. Researchers need to specify or decompose these human factors into different key variables that can be measured during the model development process. Given a deep understanding of the task relationship between human and AI, researchers can then ask further research questions on how to apply an explainable approach, or employ a HITL system vs. automated solutions, to conduct fact-checking. Here we list out several specific research topics that contain mixed-initiative tasks, including: a) assessing claim difficulty leveraging crowd workers, b) breaking down a claim into a multi-hop reasoning task and engaging the crowd to find information relevant to the sub-claims, and c) designing micro-tasks to parse a large number of documents retrieved by web search to identify sources that contain the evidence needed for veracity prediction.

\subsection{Human-centered Evaluation of NLP Technology for Fact-Checkers} \label{subsec:benchmark}


We begin this section by \replaced{proposing}{highlighting} key metrics from human factors for evaluating systems (i.e., what to measure and how to measure them): accuracy, time, model understanding, and trust (Section \ref{sub:metrics}). Following this, 
we further propose a template for an experimental protocol for human-centered evaluations in fact-checking (Section~\ref{sub:protocol}). 

\subsubsection{Metrics} \label{sub:metrics}

\paragraph{Accuracy} Most fact-checking user studies assume task accuracy as the primary user goal \citep{nguyen2018believe,mohseni2021machine}. Whereas non-expert users (i.e., social media users or other form of content consumers) might be most interested in the veracity outcome along with justification, fact-checkers often want to use automation and manual effort interchangeably in their workflow \citep{ArnoldPh:online,Nakov2021AutomatedFF}. Thus, we need a more fine-grained approach towards measuring accuracy beyond the final veracity prediction accuracy. For fine-grained accuracy evaluation, it is also crucial to capture fact-checker accuracy, particularly for the sub-tasks for which they use the fact-checking tool.

With the assumption that ``ground truth'' exists for all of the sub-tasks in the fact-checking pipeline, accuracy can be computed by comparing user answers with the ground truth. Note that measuring sub-task level accuracy is trickier than end-to-end fact-checking accuracy. Sub-task level accuracy can be captured by conducting separate experiments for each sub-task. Suppose the point of interest is to understand user performance for detecting {\em claim-checkworthiness}. In that case, we will need to collect additional data specific to the {\em claim-checkworthiness} task. 

In some cases, it is possible to merge multiple sub-tasks for evaluation purposes. For example, \citet{miranda2019automated} evaluate the effectiveness of their tool with journalists by capturing the following two key variables: a) the relevance of retrieved evidence, and b) the accuracy of the predicted stance. This method provides essential insight into evidence retrieval, stance detection, and the final fact-checking task. Depending on the tool, the exact detail of this metric will require specific changes according to tool affordances.


Note that both time and accuracy measures need to control for claim properties. For example, if a claim has been previously fact-checked, it would take less time to fact-check such claims. On the other hand, a new claim that is more difficult to assess would require more time. 

\paragraph{Model Understanding} Fact-checkers want to understand the tools they use. For example, \citet{ArnoldPh:online} pointed out that fact-checkers expressed a need for understanding CrowdTangle's algorithm for detecting viral content on various social media platforms. Similarly, \citet{Nakov2021AutomatedFF} observed a need for increased system transparency in the fact-checking tools used by different organizations. \citet{lease2018fact} argues that transparency is equally important for non-expert users to understand the underlying system and make an informed judgement. Although this is not a key variable related to user performance, it is important for practical adoption. 

To measure understanding, users could be asked to self-report their level of understanding on a Likert-scale. However, simply asking participants if they understand the algorithm is not a sufficient metric. For example, it does not indicate whether participants will be able to simulate 
tool behavior \citep{Hase2020EvaluatingEA}. We suggest the following steps for measuring model understanding based on prior work \citep{cheng2019explaining}. 

\begin{enumerate}
    \item {\bf Decision Prediction.} To capture users' holistic understanding of a tool, users could be provided claims and asked the following: ``What label would the tool assign to this claim?''
    \item {\bf Alternative Prediction.}  Capturing how changes in the input influence the output can also measure understanding, e.g., by asking users how the tool would assign a label to a claim when input parts are changed. Imagine a tool that showed the users the evidence it has considered to arrive at a veracity conclusion. Now, if certain pieces of evidence were swapped, how would that be reflected in the model prediction? 
\end{enumerate}

\paragraph{Trust} For practical adoption, trust in a fact-checking tool is crucial across all user groups. While model understanding is often positively correlated with trust, understanding alone may not suffice to establish trust. 
\deleted{and users may trust a model for even without it based on other properties of it.}
In this domain, fact-checkers and journalists may have less trust in algorithmic tools \citep{ArnoldPh:online}. On the other hand, there is also the risk of over-trust, or users blindly following model predictions \citep{nguyen2018believe,mohseni2021machine}. To maximize the tool effectiveness, we would want users to neither dismiss all model predictions out of hand (complete skepticism) nor blindly follow all model predictions (complete faith). Instead, it is important to calibrate user trust for the most effective tool usage. We suggest measuring a notion of {\em calibrated trust} \citep{Lee2004-kt}: how often users abide by correct model decisions and override erroneous model decisions.   

\begin{figure}
    \centering
    \includegraphics[width=0.4\linewidth]{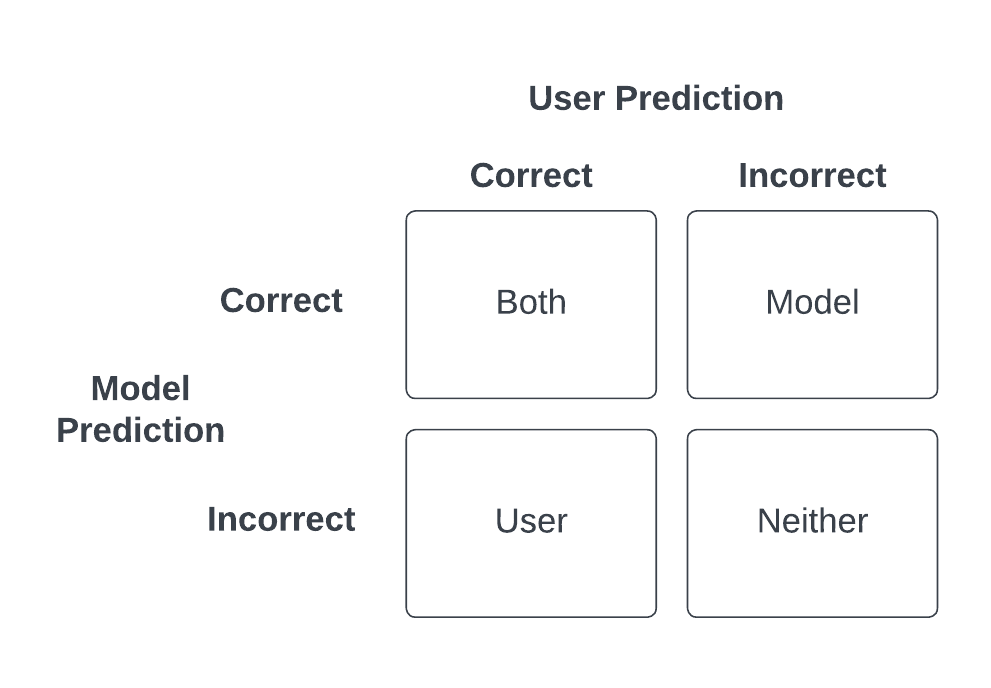}
    \caption{Confusion Matrix for User Predictions vs.\ Model Predictions with respect to ground truth (gold). 
    We assume model predictions are provided to the user, who then decides whether to accept or reject the model prediction.
    The top-left quadrant (\textit{Both}) covers cases where users  correctly follow model predictions. The top-right quadrant (\textit{Model}) denotes the cases where the model is correct but users mistakenly reject the model decisions. The bottom-left quadrant \textit{User} denotes the cases where users correctly reject erroneous model predictions. The bottom-right quadrant \textit{Neither} denotes the cases where users incorrectly accept erroneous model predictions. Quantifying user vs.\ model predictions in this manner enables measurement of {\em calibrated trust}: how often users abide by correct model decisions and override erroneous model decisions.}
    \label{tab:cm}
\vspace{-1em}
\end{figure}

To measure calibrated trust, we imagine a confusion matrix shown in Figure \ref{tab:cm}. The rows denote correct vs.\ incorrect model predictions while the columns denote correct vs.\ incorrect user predictions. A user who blindly followed all model predictions would have their behavior entirely captured by the main (primary) diagonal, whereas a user who skeptically rejected all model predictions would have their behavior captured entirely in the secondary diagonal. The ideal user's behavior would be entirely captured in the first column: accepting all correct model predictions and rejecting all incorrect model predictions. To promote effective human-AI teaming, AI tools should assist their human users in developing strong calibrated trust to appropriately trust and distrust model predictions as each case merits. 


Beyond calibrated trust, one could also measure quantitative trust by adopting methodologies from the human-machine trust literature \citep{lee1992trust}. For example, \citet{cheng2019explaining} adapted prior work into a 7-point Likert scale. A similar scale can be reused for evaluating trust in a fact-checking tool. For example, \replaced{we can create five different Likert-scales to measure the agreement (or disagreement) of users with the following statements:}{users might be asked to denote their agreement with the following statements on a Likert-scale:}
\begin{itemize}
    \item I understand the fact-checking tool. 
    \item I can predict how the tool will behave. 
    \item I have faith that the tool would be able to cope with the different fact-checking task.
    \item I trust the decisions made by the tool. 
    \item I can count on the tool to provide reliable fact-checking decisions. 
\end{itemize}

\paragraph{Additional factors} Individual differences among users might result in substantial variation in experimental outcomes. For example, varying technical literacy \citep{cheng2019explaining}, any prior knowledge about the claims, and users' political leaning \citep{thornhill2019digital} might influence user performance on the task while using fact-checking tools. Thus it is valuable to capture these factors in study design. For example:

\begin{enumerate}
    \item {\bf Technical Literacy:} Users' familiarity with popular technology tools (e.g, recommendation engines, spam detectors) and their programming experience \citep{cheng2019explaining} as well as familiarity with existing fact-checking tools. 
    \item {\bf Media Literacy:} Users' familiarity with 1) the fact-checking process, and 2) fact-checks from popular organizations such as PolitiFact and FactCheck.Org.
    \item {\bf Demographics:} Users' education level, gender, age, and political leaning.
\end{enumerate}

Quantitative measures alone are not sufficient as they do not capture certain nuances about how effectively a tool integrates into a fact-checker's workflow. For example, even if users understand and trust the working principle of a tool, it is unclear {\em why} they do so. Hence, users might be asked a few open-ended questions at the end of the study to gather qualitative insights. Such questions could include:
\begin{enumerate}
    \item Describe your understanding of the tool. Do any specific aspects of its design seem to assist or detract from your understanding of how it works? 
    \item Why do you trust or not trust the tool?
    \item Would you use this tool beyond this study, and if so, in what capacity?
\end{enumerate}










\subsubsection{Experimental Protocol} \label{sub:protocol}
One strategy to capture the aforementioned metrics is to design a mixed-methods study. Here we outline the template for such a study. Imagine the goal were to measure the user performance for fact-checking using a new tool (let's call it \textit{tool A}) compared to an existing tool (\textit{tool B}). 
Fact-checking tasks in the real world might be influenced by user priors about the claims being checked. Thus, a \textit{within-subject} study protocol may be more appropriate to account for such priors \citep{Shi-chiir22}. 

\begin{enumerate}
    \item {\bf Pre-task}: Users would first be asked to fact-check a set of claims. To do so, first a user would be asked to leverage a pre-existing \textit{tool B} at this stage. Tool B can be replaced with different baselines, depending on the particular use case, ranging from simple web-search by non-expert users to proprietary tools used by fact-checkers and journalists. Users would be asked to think aloud at this stage. 
    \item {\bf Learning}: At this stage users would familiarize themselves with the new tool (\textit{tool A}). Users would need to fact-check a different set of claims from the first one. Ground truth would also accessible to the user to form a prior about what kind of mistakes a tool might make. Claims here would be selected at random to reflect tool capabilities. Moreover, tool performance metrics would be given to the users as additional information. Users would be encouraged to ask questions about the tool at this stage.
    \item {\bf Prediction:} 
    Users would now be asked to fact-check the same claims from step-1 above but this time they are asked to leverage the \textit{tool A}. Users would be asked to think out loud through this stage. Users could simply guess the answers and achieve a high accuracy score. Thus, claims selected for stages (1) \& (3) would be a balanced set of claims with an equal distribution of true positive, true negative, false positive, and false negative samples. This idea is adopted from prior work \citep{Hase2020EvaluatingEA}.
    \item {\bf Post-task survey}: Users would now be asked to take a small survey for capturing trust, understanding, technical literacy, media literacy, and demographic information. 
    \item {\bf Post-task interview}: Upon completion of these steps, users would be interviewed with open-ended questions to gather insights about their understanding and trust in the system. 
\end{enumerate}

\noindent The measures and study protocol could be useful in the context of evaluating any new fact-checking system compared to an existing system or practices. Specifics might vary depending on the target user group and the tool's intended purpose. \added{Above we use the whole fact-checking pipeline to illustrate our experimental protocol. However this technique can be applied to other sub-tasks of automated fact checking, granted that we have the ground truth of the outcome for that sub-task.}  For example, let us assume a new claim detection tool has been proposed that takes claims from a tip-line \citep{Kazemi2021TiplinesTC}. Currently, fact-checkers use an existing claim-matching algorithm to filter out the already fact-checked claim. Now, if we replace {\em tool B} above with the existing claim-matching algorithm and {\em tool A} with the proposed claim detection tool, we can utilize the protocol mentioned above. In conclusion, one could evaluate how users perform for claim detection tasks using the new tool compared to the existing ones in terms of their accuracy, time, understanding, and trust. 

\added{
While we have proposed an ideal, extensive version of an evaluation protocol for evaluating new fact-checking tools, note that the actual protocol used in practice could be tailored according to the time required from the participants and the cost of conducting the experiment.}

\section{Conclusion} \label{sec:con}

This review highlights the practices and development of the state-of-the-art in using NLP for automated fact-checking, emphasizing both the advances and the limitations of existing task formulation, dataset construction, and modeling approaches. We partially discuss existing practices of applying these NLP tasks into real-world tools that assist human fact-checkers. In recent years we have seen significant progress in automated fact-checking using NLP. A broad range of tasks, datasets, and modeling approaches have been introduced in different parts of the fact-checking pipeline. Moreover, with recent developments in transformers and large language models, the model accuracy has improved across tasks. However, even state-of-the-art models on existing benchmarks --- such as FEVER and {\em \replaced{CLEF!}{CheckThat!}} --- may not yet be ready for practical adoption and deployment.

To address these limitations, we advocate development of hybrid, HITL systems for fact-checking. As a starting point, we may wish to reorient the goals of existing NLP tasks from full automation towards decision support. 
In contrast with fully-automated systems, hybrid systems instead involve humans-in-the-loop and facilitate human-AI teaming \citep{bansal2019beyond,bansal2021does,bansal2021most}. Such use of hybrid systems can help a)~scale-up human decision making; b)~augment machine learning capabilities with human accuracy; and c)~mitigate unintended consequences from machine errors.  Additionally, we need new benchmarks and evaluation practices that can measure how automated and hybrid systems can improve downstream human accuracy \citep{SmerosEtAl2021,fan2020generating} and efficiency in fact-checking.

\section*{Acknowledgements}

This research was supported in part by the Knight Foundation, the Micron Foundation, Wipro, and by Good Systems\footnote{\url{http://goodsystems.utexas.edu/}}, a UT Austin Grand Challenge to develop responsible AI technologies. The statements made herein are solely the opinions of the authors and do not reflect the views of the sponsoring agencies.

\bibliographystyle{cas-model2-names}

\bibliography{references}

\end{document}